\documentclass[letterpaper, 10 pt, conference]{ieeeconf}

\usepackage{amsmath}
\usepackage{amsfonts}

\usepackage{amsthm}
\usepackage[ruled,vlined]{algorithm2e}
\usepackage{mathtools}
\usepackage{tabularx}
\usepackage{graphicx}
\usepackage{subfigure}
\usepackage{enumerate}
\usepackage{float}
\usepackage{url}
\usepackage{verbatim}
\usepackage[linkcolor=black,citecolor=black,urlcolor=black,colorlinks=true]{hyperref}

\IEEEoverridecommandlockouts
\DeclareMathOperator*{\argmin}{arg\,min}
\DeclareMathOperator*{\trace}{Tr}
\newcommand{\tp}{^{\mathrm{T}}}
\newcommand{\invtp}{^{-\mathrm{T}}}
\newcommand{\grad}{\nabla}

\newcommand{\rbrac}[1]{({#1})}
\newcommand{\rBrac}[1]{\left({#1}\right)}

\newcommand{\cBrac}[1]{\left\{{#1}\right\}}
\newcommand{\norm}[1]{\Vert{#1}\Vert}
\newcommand{\Norm}[1]{\left\Vert{#1}\right\Vert}
\newcommand{\abs}[1]{\vert{#1}\vert}

\newcommand{\ceil}[1]{\lceil{#1}\rceil}

\newtheorem{theorem}{Theorem}

\title{Alternating Minimization Based Trajectory Generation for \\ Quadrotor Aggressive Flight}
\author{Zhepei Wang, Xin Zhou, Chao Xu, Jian Chu, and Fei Gao\thanks{All authors are with the State Key Laboratory of Industrial Control Technology, Institute of Cyber-Systems and Control, Zhejiang University, Hangzhou, 310027, China. {\tt\small \{wangzhepei, iszhouxin, cxu, chuj, and fgaoaa\}@zju.edu.cn}}}

\begin{document}

    \maketitle
    \thispagestyle{empty}
    \pagestyle{empty}

\begin{abstract}
With much research has been conducted into trajectory planning for quadrotors, planning with spatial and temporal optimal trajectories in real-time is still challenging. In this paper, we propose a framework for generating large-scale piecewise polynomial trajectories for aggressive autonomous flights, with highlights on its superior computational efficiency and simultaneous spatial-temporal optimality. Exploiting the implicitly decoupled structure of the planning problem, we conduct alternating minimization between boundary conditions and time durations of trajectory pieces. In each minimization phase, we leverage the algebraic convenience of the sub-problem to escape poor local minima and achieve the lowest time consumption. Theoretical analysis for the global/local convergence rate of our proposed method is provided. Moreover, based on polynomial theory, an extremely fast feasibility check method is designed for various kinds of constraints. By incorporating the method into our alternating structure, a constrained minimization algorithm is constructed to optimize trajectories on the premise of feasibility. Benchmark evaluation shows that our algorithm outperforms state-of-the-art methods regarding efficiency, optimality, and scalability. Aggressive flight experiments in a limited space with dense obstacles are presented to demonstrate the performance of the proposed algorithm. We release our implementation as an open-source ros-package\footnote{Open-source implementation is available at: \url{https://github.com/ZJU-FAST-Lab/am_traj}}.
\end{abstract}

\section{Introduction}
\label{sec:Introduction}
Recently, our community has witnessed the development of planning methods for quadrotors.
Spline-based methods, which decompose the spatial and temporal parameters of a planning problem and focus on optimizing its spatial part, are widely applied for real-time applications~\cite{Mellinger2011MinimumST,Richter2013PolynomialTP,Burri2015RealtimeVM}.

Although spline-based methods can efficiently and accurately generate energy-optimal solutions for online usage, they usually omit temporal planning for simplicity.
A typical spatial-temporal joint planning problem has high nonlinearity and nonconvexity coming from its objective and constraints.
Since temporal planning has underlying coupling with spatial parameters and implicit gradients, the spatial-temporal joint optimization cannot be solved by general nonlinear programming (NLP) in real-time.
Even though existing methods can provide online motion planning without temporal planning, they are often too conservative to be used for autonomous flights with high aggressiveness.
To bridge this gap, we propose a framework to split the spatial and temporal aspects of a trajectory optimization problem, then solve them alternately. With our method, we can obtain the energy-time joint optimal trajectory in milliseconds.

\begin{figure}[t]
    \centering
    \includegraphics[width=0.9\columnwidth]{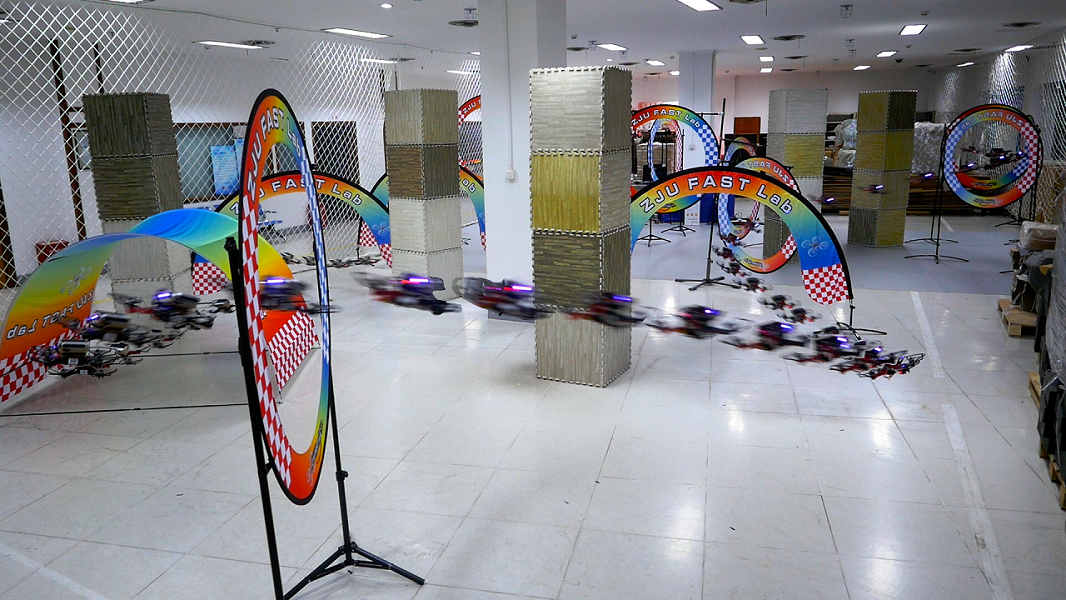}
    \caption{Composite image of the quadrotor aggressive flight in a limited space. Our quadrotor is equipped with a stereo camera, an IMU and an onboard computer. No external positioning system is used. Video of the experiment can be viewed at \url{https://youtu.be/ayoQ7i1Lz5s}. \label{fig:SequentialInstants}}
    \vspace{-0.5cm}
\end{figure}

The proposed method is based on the naturally alternating structure of the spatial-temporal trajectory optimization and designed to have guaranteed optimality and efficiency.
Our method is motivated by the fact that a polynomial trajectory with an odd order can be uniquely determined by its endpoint derivatives, i.e., the boundary condition, and the time duration. For a piecewise polynomial, once all boundary conditions are fixed, each piece of the trajectory depends only on its time duration, which can be optimized separately.
By utilizing the widely-adopted linear-quadratic objective \cite{Bertsekas1995DynamicPOC} of the optimization, the optimal time durations can be updated efficiently.
Moreover, inspired by~\cite{Richter2013PolynomialTP}, the closed-form solution is adopted to update derivatives on waypoints.
Based on the above observations, the joint optimization can be efficiently processed by an alternating minimization (AM) procedure \cite{Beck2017FOMO}. With our method, a large-scale joint optimization can be done in a few milliseconds to generate optimal trajectory with the best time allocation.

To the best knowledge of us, the proposed method is the first one that generates trajectories for a quadrotor with the spatial-temporal optimality, in such a limited time. Summarizing our contributions in this work:
\begin{itemize}
    \item An unconstrained alternating minimization algorithm is proposed to generate spatial-temporal optimal trajectories efficiently, with a proven non-asymptotic rate of the global/local convergence.
    \item A computationally efficient feasibility check method is designed for a wide range of constraints in our method.
    \item A constrained alternating minimization algorithm is constructed to optimize feasible trajectories in a recursive fashion, with global convergence verified.
    \item The proposed method is integrated into an autonomous quadrotor system then evaluated by real-world experiments as well as extensive benchmarks. The source code is released for the reference of the community.
\end{itemize}

In what follows, we discuss related literature in Sec.~\ref{sec:RelatedWork}.
Preliminaries of this paper are given in Sec.~\ref{sec:Preliminaries}.
The proposed spatial-temporal trajectory generation method for unconstrained and constrained planning cases are detailed in Sec.~\ref{sec:UnconstrainedSpatialTemporalTrajectoryOptimization} and~\ref{sec:ConstrainedSpatialTemporalTrajectoryOptimization}, respectively.
Experiments and benchmarks are given in Sec.~\ref{sec:Results}.
The paper is concluded in Sec.~\ref{sec:Conclusion}.

\section{Related Work}
\label{sec:RelatedWork}
For quadrotor planning, polynomial splines have long been used for trajectory parametrization since~\cite{Mellinger2011MinimumST}, because of their flexibility and analytical convenience.
In~\cite{Mellinger2011MinimumST}, the minimization of squared derivatives is used as the objective of quadratic programming (QP), which can be solved efficiently and accurately.
Based on this formulation, intensive works have been recently proposed. A method for obtaining a closed-form solution of the above QP program is proposed in~\cite{Richter2013PolynomialTP}, where a safe geometric path guides the generation of the trajectory to ensure its safety. By recursively adding intermediate waypoints to the path, a safe trajectory is finally generated after solving the minimum-snap problem several times. In~\cite{Liu2017PlanningDF,Gao2018OnlineSTG,Campos2017HybridOTP}, safe and dynamically feasible trajectories are online generated within a safe flight corridor, which excludes all obstacles in complex environments.
However, in these methods, the time allocation of the piece-wise trajectory is pre-defined or online adjusted by heuristics.
Although these heuristics are cheap to compute, the trajectories generated are often far less optimal and over-conservative, making them incapable of high-speed flights.

To address the time allocation problem, Mellinger et al. \cite{Mellinger2011MinimumST} compute the projected gradient with respect to durations on a hyperplane where the sum is fixed. They optimize time allocation through backtracking gradient descent. Temporal scaling is applied on the solution until dynamical feasibility is achieved. Both the finite difference and scaling used in this method are expensive operations when the number of trajectory pieces is large. Liu et al. \cite{Liu2017PlanningDF} propose a way to calculate the proper scaling factor such that a single scaling operation suffices, while it only applies for rest-to-rest trajectories. Sun et al. \cite{Sun2018FastUT} formulate the problem as a two-level optimization. They estimate the projected gradient analytically through the dual solution of the low-level QP, which improves accuracy compared with the numerical gradient. Nonetheless, the projected gradient is still inconvenient to compute and inefficient for nonlinear optimization. To avoid this situation, Richter et al. \cite{Bry2015AggressiveFO} use total duration as a regularization term, thus making each duration an independent variable. The time allocation is optimized through gradient descent, while actuator constraints are also fulfilled by scaling. However, the optimal ratio of time allocation under the constrained case may differ a lot from the unconstrained case. Consequently, scaling can ruin a trajectory where constraints are violated on a very short piece. Burri et al. \cite{Burri2015RealtimeVM} optimize the squared total duration instead. They soften all constraints by penalizing them in the objective and optimize durations through an NLP solver, while the pieces number is limited.

Gradient-based direct optimization is not satisfactory when computational overhead or solution quality is critical. To address this, Gao et al. \cite{Gao2018OptimalTA} propose a method which decouples geometrical and temporal information of a trajectory.
They use QP to generate a spatial trajectory with guaranteed safety in a virtual domain. A temporal trajectory is then optimized through second-order conic programming (SOCP) based on direct collocation \cite{Verscheure2009TimeOPT}, which maps time to the virtual domain.
Their method achieves near real-time.
The drawback is that the spatial trajectory can restrict the aggressiveness in consequent optimization, resulting in slow motions.
In \cite{Gao2019TeachRepeatReplanAC}, they improve the above method by alternating minimization between coefficients of the two-layer parametrization. Time durations of the optimized temporal trajectory is used to re-generate a spatial trajectory, on which the optimization is applied again. Although high-quality trajectories can be obtained by this process, several rounds of QP and SOCP make it unapplicable to use online.
Moreover, this method lacks theoretical convergence analysis and only works in the rest-to-rest case.
Almeida et al. \cite{Almeida2019RealTimeMS} propose a machine learning method to train a supervised neural network offline, thus online application is able to refine good initial guesses in real-time.
However, the neural network has to be trained case by case.

In this paper, we adopt the time regularized objective \cite{Bry2015AggressiveFO} as well as the alternating minimization framework \cite{Gao2019TeachRepeatReplanAC}.
For efficiency, the spatial and temporal parameters are alternately updated in separate phases.
Each minimization phase exploits the objective structure and is solved algebraically, making it free from gradient estimation and step-size choosing.
To handle various constraints, we also design a simple yet solid feasibility checker.
The proposed framework is able to generate aggressive trajectories at extremely high frequency and not limited to the rest-to-rest case.

\section{Preliminaries}
\label{sec:Preliminaries}
Differential flatness of quadrotor dynamics is validated by Mellinger et al. \cite{Mellinger2011MinimumST}.
It means the trajectory planning for a quadrotor can be done solely in the translational space. The kinodynamic feasibility is implicitly transformed into smoothness of the trajectory.
Then, actuator constraints can be enforced by restricting norms on high-order derivatives.

In this paper, we employ the piece-wise polynomial trajectory, with each piece denoted as an $N$-order polynomial:
\begin{equation}
p(t)=\mathbf{c}\tp \beta(t),~t\in[0, T],
\end{equation}
where $\mathbf{c}\in\mathbb{R}^{(N+1)\times{3}}$ is the coefficient matrix, $T$ is the duration and $\beta(t)=\rbrac{1, t, t^2, \cdots, t^N}\tp$ is a basis function.

It is worth noting that we only consider odd-order polynomial trajectories.
Since $N$ is odd, the mapping is bijective between the coefficient matrix and the boundary condition. To further explain this, consider derivatives of $p(t)$ up to $\ceil{(N-1)/2}$ order:
\begin{equation}
\mathbf{d}(t)=\rbrac{p(t), \dot{p}(t), \cdots, p^{(\ceil{(N-1)/2})}(t)}\tp,
\end{equation}
we have $\mathbf{d}(t)=\mathbf{B}(t)\mathbf{c}$ where
\begin{equation}
\mathbf{B}(t)=\rbrac{\beta(t), \dot{\beta}(t), \cdots, \beta^{(\ceil{(N-1)/2})}(t)}\tp.
\end{equation}
We denote $\mathbf{d}_{start}$ and $\mathbf{d}_{end}$ by $\mathbf{d}(0)$ and $\mathbf{d}(T)$, respectively.
The boundary condition of this polynomial is described by the tuple $\rbrac{\mathbf{d}\tp_{start}, \mathbf{d}\tp_{end}}\tp$.
The following mapping holds:
\begin{equation}
\label{eq:RepresentationMapping}
\rbrac{\mathbf{d}\tp_{start}, \mathbf{d}\tp_{end}}\tp=\mathbf{A}(T)\mathbf{c},
\end{equation}
where $\mathbf{A}(T)=\rbrac{\mathbf{B}\tp(0),\mathbf{B}\tp(T)}\tp$ is the mapping matrix.
$\mathbf{A}(T)$ is a non-singular square matrix only if $N$ is an odd number. Otherwise, $\mathbf{A}(T)$ becomes over-determined, which means for any given $\mathbf{d}_{start}$ and $\mathbf{d}_{end}$, there may not exist polynomial whose $\mathbf{c}$ satisfies (\ref{eq:RepresentationMapping}).

\begin{figure}[t]
    \centering
    \includegraphics[width=1.0\columnwidth]{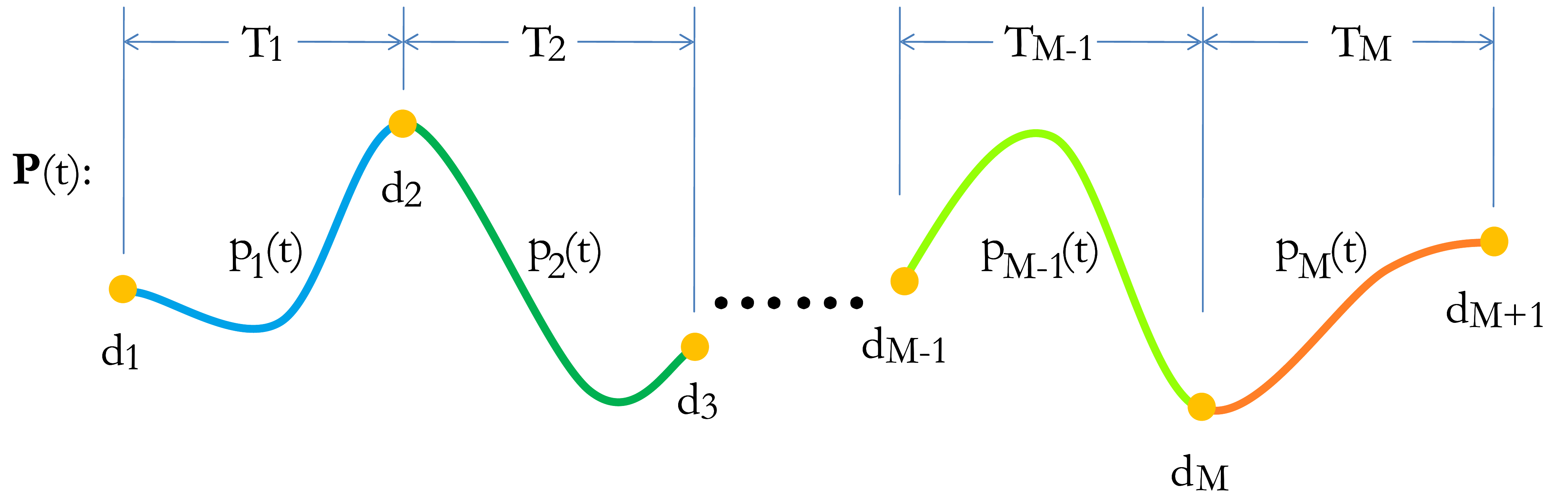}
    \vspace{-0.7cm}
    \caption{A trajectory $\mathbf{P}(t)$ contains $M$ pieces. Each piece is fully determined by its duration $T_m$ and boundary condition $\mathbf{d}_m=\rbrac{d\tp_{m}, d\tp_{m+1}}\tp$.\label{fig:TrajDiagram}}
    \vspace{-0.5cm}
\end{figure}

Moreover, the inverse $\mathbf{A}(T)^{-1}$ can be implemented with zero overhead when $N$ is an odd number. Burri et al. \cite{Burri2015RealtimeVM} explore the structure of $\mathbf{A}(T)$ and find that $\mathbf{A}(T)^{-1}$ can be computed more efficiently through its Schur-Complement, which only involves submatrix inverse. We take things one step further. Actually, all entries of $\mathbf{A}(T)^{-1}$ are power functions of $T$, thus Gaussian-Elimination is applied to get its analytic form. Consequently, time-consuming operation is no longer needed when $\mathbf{A}(T)^{-1}$ is computed online. To achieve this, we pre-compute matrices $\mathbf{E}, \mathbf{F}, \mathbf{G}, \mathbf{U}, \mathbf{V}, \mathbf{W} \in \mathbb{R}^{S\times{S}}$ offline, where $S=(N+1)/2$:
\[
\mathbf{E}_{ij}=
\begin{cases}
\prod\limits_{k=1}^{i-1}{k} & \text{if } i=j,\\
0 & \text{if } i\neq{j}.
\end{cases},~
\mathbf{F}_{ij}=
\begin{cases}
\prod\limits_{k=j-i+1}^{j-1}{k} & \text{if } i\leq{j},\\
0 & \text{if } i>j.
\end{cases}
\]
\[
\mathbf{G}_{ij}=\prod\limits_{k=S-i+j+1}^{S+j-1}{k},~
\mathbf{U}_{ij}=
\begin{cases}
1/\prod\limits_{k=1}^{i-1}{k} & \text{if } i=j,\\
0 & \text{if } i\neq{j}.
\end{cases}
\]
\[
\mathbf{W}=\mathbf{G}^{-1},\mathbf{V}=-\mathbf{W}\mathbf{F}\mathbf{U}.
\]
Following mapping matrices are computed online:
\begin{equation*}
\mathbf{A}(T)=\begin{pmatrix}\mathbf{E}&\mathbf{0}\\\rBrac{\mathbf{F}_{ij}T^{j-i}}_{S\times{S}}&\rBrac{\mathbf{G}_{ij}T^{S-i+j}}_{S\times{S}}\end{pmatrix},
\end{equation*}
\begin{equation*}
\mathbf{A}(T)^{-1}=\begin{pmatrix}\mathbf{U}&\mathbf{0}\\\rBrac{\mathbf{V}_{ij}T^{j-i-S}}_{S\times{S}}&\rBrac{\mathbf{W}_{ij}T^{j-i-S}}_{S\times{S}}\end{pmatrix}.
\end{equation*}
Therefore, provided with an odd order, we show the practical equivalence between the tuple $\rbrac{\mathbf{d}_{start}, \mathbf{d}_{end}, T}$ and $\rbrac{\mathbf{c}, T}$ in the sense of polynomial representation.

Consequently, we consider an $M$-piece trajectory $\mathbf{P}$ parametrized by time allocation $\mathbf{T}=\rbrac{T_1, T_2, \cdots, T_M}\tp$ as well as boundary conditions $\mathbf{D}=\rbrac{d\tp_1, d\tp_2, \cdots, d\tp_{M+1}}\tp$ of all pieces, as shown in Fig.~\ref{fig:TrajDiagram}. The trajectory is defined by
\begin{equation}
\mathbf{P}(t):=\mathbf{d}_m\tp\mathbf{A}(T_m)\invtp\beta( t-\sum_{i=1}^{m-1}{T_i}),
\end{equation}
where $t$ lies in the $m$-th piece and $\mathbf{d}_m=\rbrac{d\tp_{m}, d\tp_{m+1}}\tp$ is a boundary condition of the $m$-th piece.
This definition implicitly involves $(N-1)/2$ order continuity at boundaries of each piece.
Normally, some entries in $\mathbf{D}$ are fixed, such as the position of waypoints \cite{Richter2013PolynomialTP}.
We split $\mathbf{D}$ into two parts, the fixed part $\mathbf{D}_F$ which is viewed as constant, and the free part $\mathbf{D}_P$ which is to be optimized.
Then, the whole trajectory can be fully determined by $\mathbf{P}=\mathbf{\Phi}(\mathbf{D}_P, \mathbf{T})$.

\section{Spatial-Temporal Trajectory Optimization: \textit{Unconstrained Case}}
\label{sec:UnconstrainedSpatialTemporalTrajectoryOptimization}
In this section, we describe our method for jointly optimizing spatial and temporal parameters of a trajectory, for the \textit{Unconstrained Case}, where no constraint is considered.

\subsection{Optimization Objective}
We use the time regularized quadratic cost over the whole trajectory, as the objective of the optimization:
\begin{equation}
J(\mathbf{P})=\int_{0}^{\sum_{m=1}^{M}{T_m}}\rBrac{{\rho + \sum_{i=D_{min}}^{D_{max}}{w_i\Norm{\mathbf{P}^{(i)}(t)}^2}}}\mathrm{d}t,
\end{equation}
where $D_{min}$ and $D_{max}$ are the lowest and the highest order of derivative to be penalized respectively, $w_i$ is the weight of the $i$-order derivative and $\rho$ is the weight of time regularization. The weight $\rho$ adjusts the aggressiveness of the trajectory \cite{Burri2015RealtimeVM}, which allows total duration varies adaptively. For now, we consider the unconstrained optimization:
\begin{equation}
\label{eq:UnconstrainedObjective}
\min_{\mathbf{D}_P, \mathbf{T}}{J(\mathbf{D}_P, \mathbf{T})}
\end{equation}
where free boundary conditions and durations are decision variables. $J(\mathbf{D}_P, \mathbf{T}):=J(\mathbf{\Phi}(\mathbf{D}_P, \mathbf{T}))$ is used for brevity.

The cost $J_m$ for the $m$-th piece can be calculated as
\begin{equation}
\label{eq:KthPieceObjective}
J_m=\rho T_m+\trace\cBrac{\mathbf{d}_m\tp\mathbf{A}(T_m)\invtp \mathbf{Q}(T_m)\mathbf{A}(T_m)^{-1}\mathbf{d}_m},
\end{equation}
in which $\mathbf{Q}(T_m)$ is a symmetric matrix \cite{Bry2015AggressiveFO} consisting of high powers of $T_m$, and $\trace\cBrac{\cdot}$ is trace operation which only sums up diagonal costs produced in three dimensions. The overall objective can be formulated as
\begin{equation}
\label{eq:ObjectiveStructure}
J=\rho\Norm{\mathbf{T}}_1+\trace\cBrac{\begin{pmatrix}\mathbf{D}_F\\\mathbf{D}_P\end{pmatrix}\tp\mathbf{C}\tp\mathbf{H}(\mathbf{T})\mathbf{C}\begin{pmatrix}\mathbf{D}_F\\\mathbf{D}_P\end{pmatrix}},
\end{equation}
\begin{equation}
\mathbf{H}(\mathbf{T})=\bigoplus_{m=1}^{M}{\mathbf{A}(T_m)\invtp\mathbf{Q}(T_m)\mathbf{A}(T_m)^{-1}},
\end{equation}
where $\mathbf{H}(\mathbf{T})$ is the direct sum of its $M$ diagonal blocks, and $\mathbf{C}$ is a permutation matrix. We make sure that the setting for $J$ is legal by assuming that the $\alpha$-sublevel set of $J(\mathbf{D}_P, \mathbf{T})$ for any finite $\alpha$  is bounded and only consists of positive time allocation. For example, consecutive repeating waypoints with identical boundary conditions fixed in $\mathbf{D}_F$ are not allowed.

\subsection{Unconstrained Trajectory Optimization}
To optimize Eq.~(\ref{eq:ObjectiveStructure}), we propose an AM-based algorithm, as shown in Alg.~\ref{alg:UnconstrainedSpatialTemporalAM}.
Initially, $\mathbf{T}^{0}$ is solved for the provided $\mathbf{D}_P^0$. After that, the minimization of the objective function is done through a two-phase process, in which only one of $\mathbf{D}_P$ and $\mathbf{T}$ is optimized while the other is fixed.

\begin{algorithm}
    \caption{Unconstrained Spatial-Temporal AM}
    \label{alg:UnconstrainedSpatialTemporalAM}
    \KwIn{$\mathbf{D}_P^0, K\in\mathbb{Z}_+, \delta>0$}
    \KwOut{$\mathbf{D}_P^*, \mathbf{T}^*$}
    \Begin
    {
        $\mathbf{T}^0 \leftarrow \argmin_{\mathbf{T}}{J(\mathbf{D}_P^0, \mathbf{T})}$\;
        $J_l \leftarrow J(\mathbf{D}_P^0, \mathbf{T}^0), k \leftarrow 0$\;
        \While{$k<K$}
        {
            $\mathbf{D}_P^{k+1} \leftarrow \argmin_{\mathbf{D}_P}{J(\mathbf{D}_P, \mathbf{T}^{k})}$\;
            $\mathbf{T}^{k+1} \leftarrow \argmin_{\mathbf{T}}{J(\mathbf{D}_P^{k+1}, \mathbf{T})}$\;
            $J_{c} \leftarrow J(\mathbf{D}_P^{k+1}, \mathbf{T}^{k+1})$\;
            \If{$\abs{J_l-J_c}<\delta$}{\textbf{break}}
            $J_l \leftarrow J_c, k \leftarrow k+1$\;
        }
    $\mathbf{D}_P^* \leftarrow \mathbf{D}_P^{k}, \mathbf{T}^* \leftarrow \mathbf{T}^{k}$\;

    \Return{$\mathbf{D}_P^*, \mathbf{T}^*$};
    }
\end{algorithm}

In the first phase, the sub-problem
\begin{equation}
\label{eq:OptimizeD}
\mathbf{D}_P^*(\mathbf{T})=\argmin_{\mathbf{D}_P}{J(\mathbf{D}_P, \mathbf{T})}
\end{equation}
is solved for each $\mathbf{T}^k$.
We employ the unconstrained QP formulation by Richter et al. \cite{Bry2015AggressiveFO}, and briefly introduce it here.
The matrix $\mathbf{R}(\mathbf{T})=\mathbf{C}\tp\mathbf{H}(\mathbf{T})\mathbf{C}$ is partitioned as
\begin{equation}
\mathbf{R}(\mathbf{T})=\begin{pmatrix}\mathbf{R}_{FF}(\mathbf{T}) & \mathbf{R}_{FP}(\mathbf{T}) \\
\mathbf{R}_{PF}(\mathbf{T}) & \mathbf{R}_{PP}(\mathbf{T})\end{pmatrix},
\end{equation}
then the solution is be obtained analytically through
\begin{equation}
\label{eq:AnalyticalD}
\mathbf{D}_P^*(\mathbf{T})=-\mathbf{R}_{PP}(\mathbf{T})^{-1}\mathbf{R}_{FP}(\mathbf{T})\mathbf{D}_F.
\end{equation}
For efficiency, we solve the sparse linear system
\begin{equation}
\mathbf{R}_{PP}(\mathbf{T})X=-\mathbf{R}_{FP}(\mathbf{T})\mathbf{D}_F
\end{equation}
through Sparse LU Factorization to get $\mathbf{D}_P^*(\mathbf{T})$ since $\mathbf{H}(\mathbf{T})$ and $\mathbf{C}$ are both sparse.

In the second phase, the sub-problem
\begin{equation}
\label{eq:OptimizeT}
\mathbf{T}^*(\mathbf{D}_P)=\argmin_{\mathbf{T}}{J(\mathbf{D}_P, \mathbf{T})}
\end{equation}
is solved for each $\mathbf{D}_P^k$. In this phase, the scale of sub-problem can be reduced into each piece.
Due to our representation of trajectory, once $\mathbf{D}_P$ is fixed, the boundary conditions $\mathbf{D}$ isolate each entry in $\mathbf{T}$ from the others. Therefore, $T_m$ can be optimized individually to get all entries of $\mathbf{T}^*(\mathbf{D}_P)$. As for the $m$-th piece, its cost $J_m$ in (\ref{eq:KthPieceObjective}) is indeed a rational function of $T_m$. We show the structure of $J_m$ and omit the deduction for brevity:
\begin{equation}
\label{eq:RationalCostOnPiece}
J_m(T)=\rho T+\frac{1}{T^{p_{n}}}\sum\limits_{i=0}^{p_{d}}{\alpha_i T^i},
\end{equation}
where $p_{n}=2D_{max}-1$ and $p_{d}=2(D_{max}-D_{min})+N-1$ are orders of numerator and denominator, respectively. The coefficient $\alpha_i$ is determined by $\mathbf{d}_m$. Due to positiveness of $J_m(T)$, we have $J_m(T)\rightarrow+\infty$ as $T\rightarrow+\infty$ or $T\rightarrow0^+$. Therefore, the minimizer exists for
\begin{equation}
T_m^*(\mathbf{D}_P)=\argmin_{T\in\rbrac{0, +\infty}}{J_m(T)}.
\end{equation}
To find all candidates, we compute the derivative of (\ref{eq:RationalCostOnPiece}):
\begin{equation}
\frac{\mathrm{d}J_m(T)}{\mathrm{d}T}=\rho+\frac{1}{T^{1+p_{n}}}\sum\limits_{i=0}^{p_{d}}{(i-p_n)\alpha_i T^i}.
\end{equation}
The minimum exists in solutions of ${\mathrm{d}J_m(T)}/{\mathrm{d}T}=0$, which can be calculated through any modern univariate polynomial real-roots solver \cite{Sagraloff2013ComputingRR}.
In this paper, we utilize the Continued Fraction method \cite{Tsigaridas2006UnivariatePR} to isolate all positive roots of any high order ($\geq5$) polynomial. The second phase is completed by updating every entry $T_m^*(\mathbf{D}_P)$ in $\mathbf{T}^*(\mathbf{D}_P)$.

\subsection{Convergence Analysis}
Alg.~\ref{alg:UnconstrainedSpatialTemporalAM} is globally convergent. Moreover, it is faster than conventional gradient descent used in time allocation refinement, under no assumption on convexity.

\begin{theorem}
    \label{thm:UnconstrainedGlobalConvergence}
    Consider the process described in Algorithm \ref{alg:UnconstrainedSpatialTemporalAM}. Provided with any $\mathbf{D}_P^0$, the following inequality always holds for the $K$-th iteration:
    \[
    \min_{0\leq{k}\leq{K}}{\norm{\grad{J(\mathbf{D}_P^k, \mathbf{T}^k)}}}_F^2\leq{M_c \frac{J(\mathbf{D}_P^0, \mathbf{T}^0)-J_c}{K}},
    \]
    where $M_c$ and $J_c$ are both constant, $\Norm{\cdot}_F$ is Frobenius norm. It shows the process globally converges to a stationary point with non-asymptotic sublinear rate $O(1/\sqrt{K})$.
\end{theorem}
\begin{proof}
    See \cite{Wang2020DetailedPAM} for details.
\end{proof}

Thm.~\ref{thm:UnconstrainedGlobalConvergence} shows that our algorithm shares the same global convergence rate as that of gradient descent with the best step-size chosen in each iteration~\cite{Nesterov2018LecturesOC}. The best step-size is practically unavailable.
In contrast, our algorithm does not involve any step-size choosing in its iterations.
Sub-problems in Eq.~(\ref{eq:OptimizeD}) and Eq.~(\ref{eq:OptimizeT}) both are solved exactly and efficiently due to their algebraic convenience. Therefore, Alg.~\ref{alg:UnconstrainedSpatialTemporalAM} is faster than gradient-based methods in practice.

Another key advantage of our algorithm is its capability of escaping from a local minimum in the time optimization.
Watching Eq.~(\ref{eq:ObjectiveStructure}), despite $J(\mathbf{D}_P, \mathbf{T})$ is convex in $\mathbf{D}_P$ as proved in \cite{Wang2020DetailedPAM}, it is a rational function which can have multiple local minima in $T_m$.
Therefore, a case may occur where the initial time allocation falls into one of these local minima instead of the global minimum in $(0, +\infty)$.
Under this situation, naturally, the global minimum in time allocation cannot be attained by gradient-based methods.
However, with our method, all local minima are compared directly. Thus, the situation can be avoided.

It is worth noting that, here the global optimality is not guaranteed because our algorithm still exploits local structure of the problem.
Although convergence to a stationary point is ensured, we argue that strict saddle points are theoretically and numerically unstable for our first-order AM method~\cite{Lee2019FirstorderMA}. Moreover, when the stationary point is a strict local minimum, we show that the convergence rate is faster.

\begin{theorem}
    \label{thm:UnconstrainedLocalConvergence}
    Let $(\widehat{\mathbf{D}}_P, \widehat{\mathbf{T}})$ denote any strict local minimum of $J(\mathbf{D}_P, \mathbf{T})$ to which Alg.~\ref{alg:UnconstrainedSpatialTemporalAM} converges. There exist $K_c\in\mathbb{Z}_+$ and $\gamma\in\mathbb{R}_+$, such that
    \[
    J(\mathbf{D}_P^K, \mathbf{T}^K)-J^*\leq{\frac{1}{\gamma(K-K_c)+(J(\mathbf{D}_P^{K_c}, \mathbf{T}^{K_c})-J^*)^{-1}}},
    \]
    for all $K\geq{K_c}$, where $J^*=J(\widehat{\mathbf{D}}_P, \widehat{\mathbf{T}})$.
\end{theorem}
\begin{proof}
    See \cite{Wang2020DetailedPAM} for details.
\end{proof}

Thm.~\ref{thm:UnconstrainedLocalConvergence} shows that a faster convergence rate $O(1/K)$ can be attained for a strict local minimum than the general case in Thm.~\ref{thm:UnconstrainedGlobalConvergence}. Note that it is possible to accelerate our method to attain the optimal rate $O(1/K^2)$ of first-order methods or use second-order methods to achieve better performance.
However, we still employ the first-order AM process because of its simplicity in implementation and its good performance when the trajectory is far from optimum.

The non-asymptotic property implies that the convergence is bounded strictly by the rate, rather than approximated. This property, along with the monotone decrease of the objective, shows guaranteed progress in each iteration, while gradient-based methods may try bad step-size thus making no/negative progress.

\section{Spatial-Temporal Trajectory Optimization: \textit{Constrained Case}}
\label{sec:ConstrainedSpatialTemporalTrajectoryOptimization}
In this section, we present our method to incorporate safety and dynamical feasibility constraints into our optimization process.
To begin with, we introduce a computationally efficient feasibility check method that applies to a wide range of constraints.
Then this method is used in a constrained trajectory optimization process.

\subsection{Computationally Efficient Feasibility Check}
\label{subsec:EfficientFeasibilityCheck}
A piece of the trajectory is denoted by
\begin{equation}
p(t)=\rbrac{p_1(t), p_2(t), p_3(t)}\tp.
\end{equation}
Constraint $G(p_1^{(i)}(t),p_2^{(i)}(t),p_3^{(i)}(t))<{0}$ over $[0,T]$ should be satisfied by the $i$-order derivative of the piece. It is required that $G$ has the form of a multivariate polynomial:
\begin{equation}
\label{eq:ConstraintForm}
G(a,b,c):=\sum_{e_1+e_2+e_3\leq{d_g}}^{d_c\in\mathbb{R}, e_j\in\mathbb{N}}{d_c\cdot{a^{e_1}b^{e_2}c^{e_3}}},
\end{equation}
where $d_g$ is the highest degree. Many kinds of constraints can be expressed by $G$, such as the safe distance constraint to keep away from an obstacle located at $ \rbrac{0,0,0}\tp$:
\begin{align*}
&G_{p}(p_1(t),p_2(t),p_3(t))<{0},\\
&G_{p}(a,b,c):=r_{safe}^2-(a^2+b^2+c^2),
\end{align*}
or maximum speed constraint:
\begin{align*}
&G_{v}(\dot{p}_1(t),\dot{p}_2(t),\dot{p}_3(t))<{0},\\
&G_{v}(a,b,c):=a^2+b^2+c^2-v^2_{max}.
\end{align*}

Provided with any piece $p(t)$, we check whether constraint $G$ is fulfilled for all $t\in[0,T]$.
We define $\mathcal{G}(t):=G(p_1^{(i)}(t),p_2^{(i)}(t),p_3^{(i)}(t))$ which is indeed a polynomial of $t$.
The procedure is as follows:
Firstly, check the sign of $\mathcal{G}(0)$ and $\mathcal{G}(T)$.
Then, If both two endpoints satisfy the constraint, we have to make sure the constraint is not violated inside the interval $(0, T)$.
Instead of locating all extrema of $\mathcal{G}(t)$ and checking their values, we only need to check the existence of root of $\mathcal{G}(t)=0$ in the interval.
If the equation has any root in $(0,T)$, then $p(t)$ is infeasible.
Fortunately, it is convenient for a polynomial to achieve this leveraging \textit{Sturm's Theory} \cite{Basu2003AlgorithmsIR}.
Now that neither $0$ nor $T$ is the root of $\mathcal{G}(t)=0$, we compute the Sturm sequence $g_0(t), g_1(t),g_2(t),\cdots$ by
\begin{align}
\label{eq:SturmSequence}
g_0(t)&=\mathcal{G}(t), \nonumber \\
g_1(t)&=\dot{\mathcal{G}}(t), \\
-g_{k+1}(t)&=Rem(g_{k-1}(t), g_k(t))  \nonumber,
\end{align}
where $Rem(g_{k-1}(t), g_k(t))$ is remainder in the Euclidean division of $g_{k-1}(t)$ by $g_k(t)$ \cite{Basu2003AlgorithmsIR}.
When $g_k(t)$ becomes constant, we stop expanding this sequence.
Let $V_{sign}(\tau)$ denote the number of sign variations of Sturm sequence at $t=\tau$, in which zero values should be ignored.
Then the number of distinct roots inside $(0, T)$ equals $V_{sign}(0)-V_{sign}(T)$.
Here the feasibility check is done for $G(\cdot,\cdot,\cdot)<{0}$.
Sometimes, constraint has the form $G(\cdot,\cdot,\cdot)\leq{0}$.
In practice, it can be equally handled by checking $G(\cdot,\cdot,\cdot)<\epsilon$, where $\epsilon$ is a small positive real number. What's more, non-polynomial constraint can also be efficiently checked through its Taylor series within acceptable approximation error.

In conclusion, our method converts the feasibility check into the root existence check for polynomial, without computing the root itself.
Our method is straightforward and involves no redundant operations, in comparison with methods used in \cite{Mueller2015ACE} and \cite{Burri2015RealtimeVM}.

\subsection{Constrained Trajectory Optimization}
For the \textit{Constrained Case}, we enforce constraints on norms of derivatives of the trajectory:
\begin{align}
\label{eq:ConstrainedObjective}
\min_{\mathbf{D}_P, \mathbf{T}}&{J(\mathbf{D}_P, \mathbf{T})}\\
\label{eq:Constraints}
s.t.~&\norm{\mathbf{P}^{(n)}(t)}\leq\sigma_n,n=1,\cdots,N\\
&\mathbf{P}=\mathbf{\Phi}(\mathbf{D}_P,\mathbf{T})
\end{align}

Generally, the constraint does not have to be like (\ref{eq:Constraints}). If only a constraint is representable in (\ref{eq:ConstraintForm}) and its feasible solution can be constructed, then it can be handled in our optimization. With a slight abuse of notation, we use $\mathbf{G}(\mathbf{D}_P, \mathbf{T})\leq0$ to denote that $\mathbf{\Phi}(\mathbf{D}_P,\mathbf{T})$ is feasible. $\mathbf{G}(\mathbf{d}_m, T_m)\leq{0}$ is used to denote that the $m$-th piece is feasible. We say $\mathbf{G}(\mathbf{d}_m, T_m)\leq{0}$ is tight by meaning that, at least one constraint is tight at a $t$ on the $m$-th piece.

\begin{figure}[t]
    \begin{center}
        \subfigure[\label{fig:ConstrainedOptimizationDp} Update of Derivatives.]
        {\includegraphics[width=0.85\columnwidth]{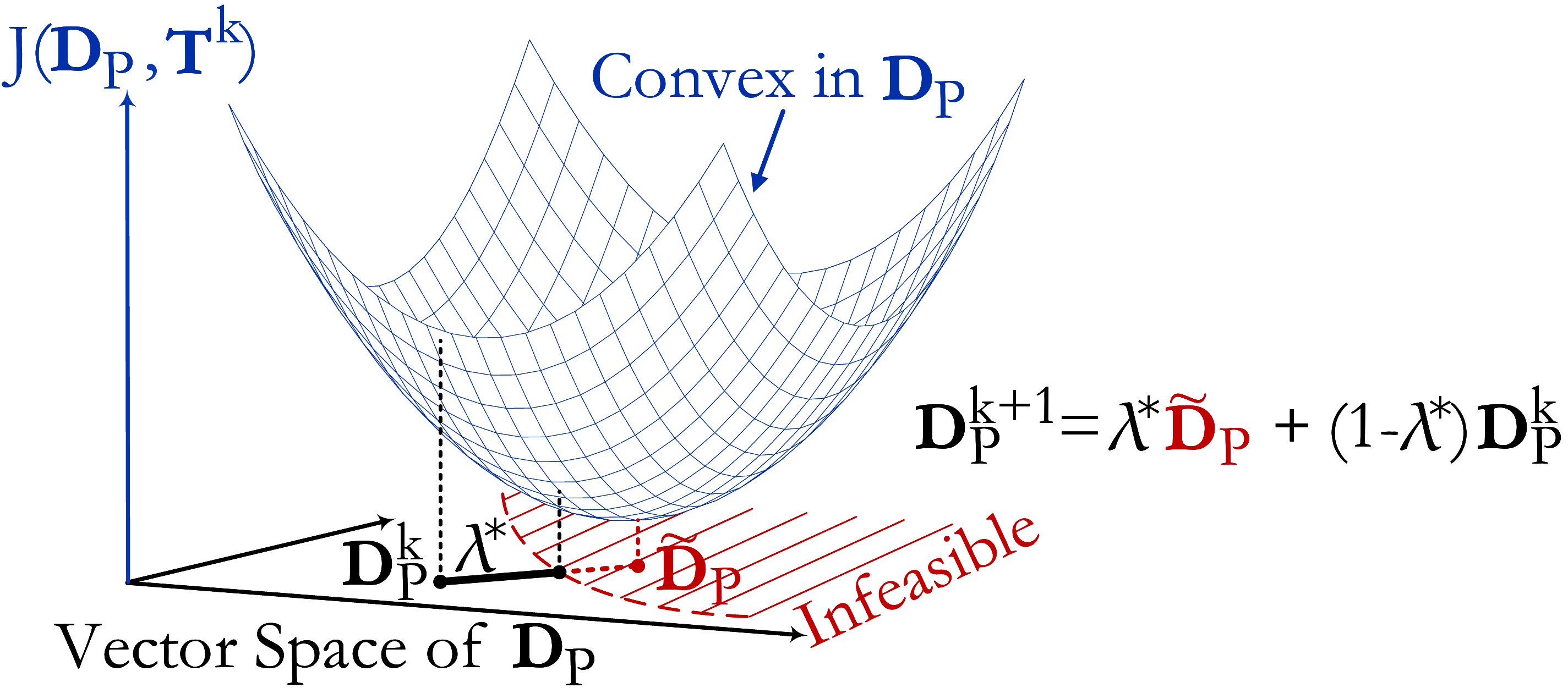}}
        \subfigure[\label{fig:ConstrainedOptimizationT} Update of time allocation.]
        {\includegraphics[width=0.6\columnwidth]{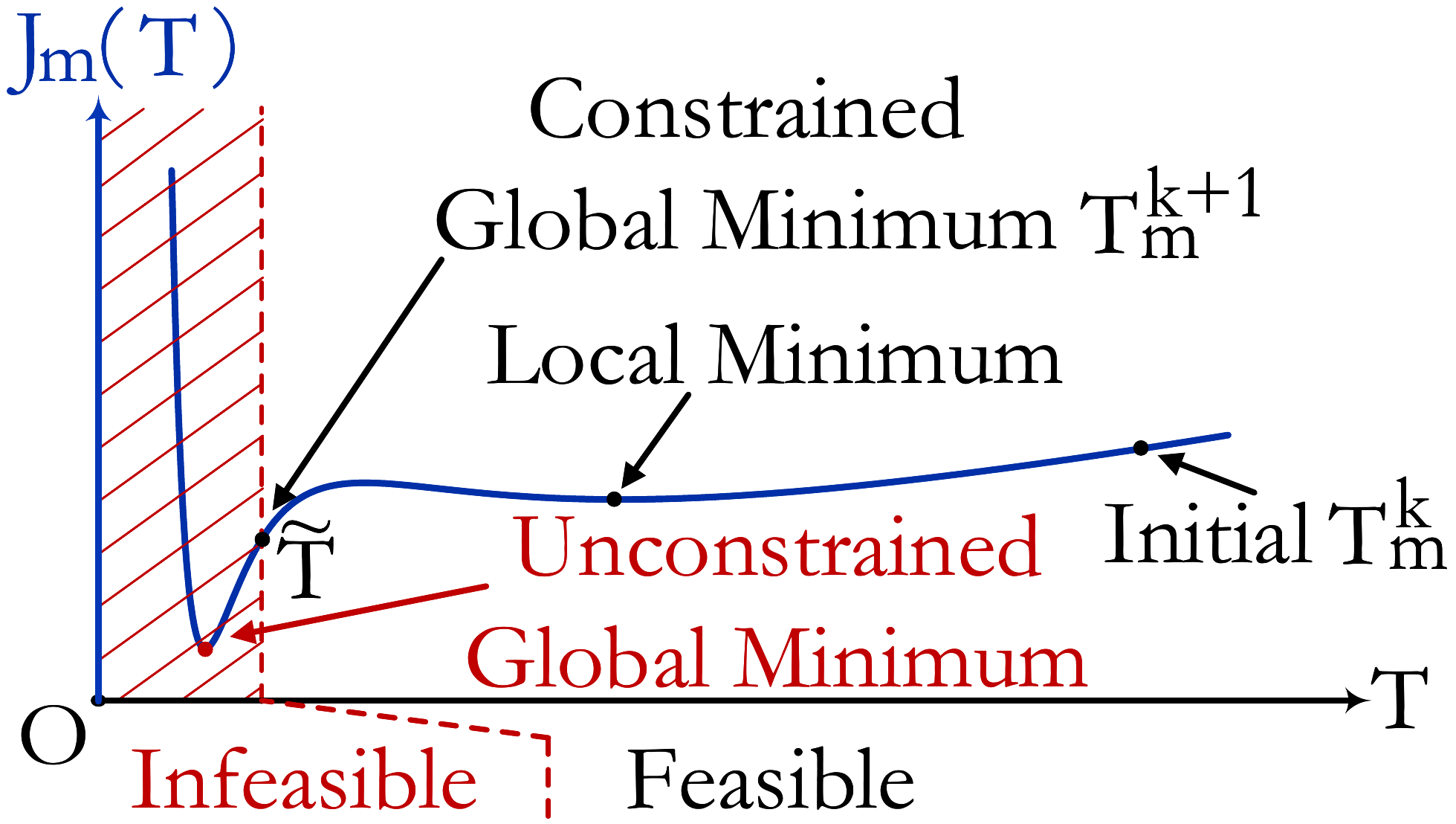}}
    \end{center}
    \vspace{-0.3cm}
    \caption{\label{fig:TwoPhaseConstrainedOptimization} Illustration for the two phases in constrained optimization. }
    \vspace{-0.5cm}
\end{figure}

The constrained version of our method is shown in Alg.~\ref{alg:ConstrainedSpatialTemporalAM}. An initial feasible trajectory $\mathbf{P}^0$ can be constructed from conservative time allocation. The spatial-temporal parameters $(\mathbf{D}_P^0, \mathbf{T}^0)$ are then recovered from the trajectory, which is used in the consequent two-phase constrained minimization.

In the first phase, $\mathbf{T}^k$ is fixed. An illustration is provided in Fig.~\ref{fig:ConstrainedOptimizationDp}, where the unconstrained minimum $\widetilde{\mathbf{D}}_P$ is obtained as is done in Alg.~\ref{alg:UnconstrainedSpatialTemporalAM}. The trajectory $\mathbf{\Phi}(\widetilde{\mathbf{D}}_P, \mathbf{T}^k)$ may not be feasible.
Since the feasibility of $(\mathbf{D}_P^k, \mathbf{T}^k)$ is ensured in last iteration, a line search is done as
\begin{equation}
\min_{\lambda\in[0,1]}{J(\mathbf{D}(\lambda), \mathbf{T}^{k})},~s.t.~\mathbf{G}(\mathbf{D}(\lambda), \mathbf{T}^k)\leq{0},
\end{equation}
where $\mathbf{D}(\lambda)$ is the convex combination of $\widetilde{\mathbf{D}}_P$ and $\mathbf{D}_P^k$. Convexity of $J(\cdot, \mathbf{T}^{k})$ implies the convexity of $J(\mathbf{D}(\cdot), \mathbf{T}^{k})$.
Moreover, $\lambda=0$ is a feasible solution, while $\lambda=1$ is the unconstrained global minimum. We simply take $\lambda^*=1$ if it is feasible. If not, a bisection procedure is done on the interval $[0,1]$. In this procedure, the feasibility check method described in Sec.~\ref{subsec:EfficientFeasibilityCheck} is employed to shrink the interval.
The procedure stops at an acceptable interval length, with $\lambda^*$ taking the feasible lower bound.
After that, we update $\mathbf{D}_P^{k+1}$ by $\mathbf{D}(\lambda^*)$.
Meanwhile, a set $\Delta$ is maintained to store indices of tightened pieces.

\begin{algorithm}
    \caption{Constrained Spatial-Temporal AM}
    \label{alg:ConstrainedSpatialTemporalAM}
    \KwIn{Feasible $\mathbf{P}^0$, $K\in\mathbb{Z}_+, \delta>0$}
    \KwOut{$\mathbf{P}^*$}
    \Begin
    {
        $(\mathbf{D}_P^0, \mathbf{T}^0)\leftarrow\mathbf{\Phi}^{-1}(\mathbf{P}^0)$\;
        $J_l \leftarrow J(\mathbf{D}_P^0, \mathbf{T}^0)$\;
        $k \leftarrow 0, \Delta \leftarrow \cBrac{}$\;
        \While{$k<K$}
        {
            $\widetilde{\mathbf{D}}_P \leftarrow \argmin_{\mathbf{D}_P}{J(\mathbf{D}_P, \mathbf{T}^{k})}$\;
            $\mathbf{D}(\lambda):=\lambda\widetilde{\mathbf{D}}_P+(1-\lambda)\mathbf{D}_P^{k}$\;
            $\lambda^* \leftarrow \argmin_{\lambda\in[0,1]}{J(\mathbf{D}(\lambda), \mathbf{T}^{k})}$\\
            $~~~~~~~~s.t.~\mathbf{G}(\mathbf{D}(\lambda), \mathbf{T}^k)\leq{0}$\;
            $\mathbf{D}_P^{k+1}=\mathbf{D}(\lambda^*)$\;
            Recover  $T^k_m$ from $\mathbf{T}^k$ and $\mathbf{d}_m^{k+1}$ from $\mathbf{D}_P^{k+1}$\;
            \For{$m\leftarrow 1$ \KwTo $M$}
            {
                \uIf{$\mathbf{G}(\mathbf{d}_m^{k+1}, T_m^k)\leq{0}$ is tight}
                {
                    $\Delta \leftarrow \Delta\cup\cBrac{m}$\;
                }
                \Else
                {
                    $\Delta \leftarrow \Delta\setminus\cBrac{m}$\;
                }
            }
            \For{$m\leftarrow 1$ \KwTo $M$}
            {
                Construct $J_m(T)$ from $\mathbf{d}_m^{k+1}$\;
                $T_m^{k+1} \leftarrow \argmin_{T\in(0,+\infty)}{J_m(T)}$\\
                $~~~~~~~~~~~s.t.~\mathbf{G}(\mathbf{d}_m^{k+1}, T)\leq{0}$\;
            }
           Construct $\mathbf{T}^{k+1}$ from $T_m^{k+1}$\;
           $J_{c} \leftarrow J(\mathbf{D}_P^{k+1}, \mathbf{T}^{k+1})$\;
           \If{$\abs{J_l-J_c}<\delta$}{\textbf{break}}
           $J_l \leftarrow J_c, k \leftarrow k+1$\;
       }
      $\mathbf{P}^k\leftarrow\mathbf{\Phi}(\mathbf{D}_P^k,\mathbf{T}^k)$\;
      \If{$\Delta$ is not empty}
      {
          Split $\mathbf{P}^k$ by removing pieces in $\Delta$\;
          Call this \textbf{Algorithm} on sub-trajectories\;
          Update sub-parts of $\mathbf{P}^k$\;
      }
  $\mathbf{P}^* \leftarrow \mathbf{P}^k$\;
  \Return{$\mathbf{P}^*$};
}
\end{algorithm}

In the second phase, $\mathbf{D}_P^{k+1}$ is fixed. An illustration is given in Fig.~\ref{fig:ConstrainedOptimizationT}. Each entry in $\mathbf{T}^k$ is updated by solving
\begin{equation}
\min_{T\in(0,+\infty)}{J_m(T)},~s.t.~\mathbf{G}(\mathbf{d}_m^{k+1}, T)\leq{0}.
\end{equation}
As stated in Alg.~\ref{alg:UnconstrainedSpatialTemporalAM}, all extrema of $J_m(T)$ can be computed exactly. However, the constrained minimum may not exist in them. It can be any $\widetilde{T}$ at which some constraints are exactly tightened. When infeasible extremum exists, $\widetilde{T}$ must be located between any infeasible extremum and the neighboring feasible one or $T_m^k$. A bisection procedure with feasibility check suffices to compute $\widetilde{T}$. After that, we compare $J_m(T)$ on those feasible extrema together with $\widetilde{T}$.

When all iterations are done, the set $\Delta$ indicates pieces stuck by active constraints. If $\Delta$ is not empty, we recursively apply Alg.~\ref{alg:ConstrainedSpatialTemporalAM} on split sub-trajectories, while boundary conditions of pieces indexed by $\Delta$ should be totally fixed. Finally, $\mathbf{P}^*$ is updated and returned. The recursive process is essential, since it ensures that pieces with no room for optimization do not prevent other pieces to decrease the objective. Alg.~\ref{alg:ConstrainedSpatialTemporalAM} is globally convergent to a solution set where constraints are tight or local minimum is attained, which can be checked by Zangwill's theorem \cite{Zangwill1972NonlinearP}.

\section{Results}
\label{sec:Results}
\subsection{Comparison of Feasibility Check Methods}
\begin{figure}[t]
    \centering
    \includegraphics[width=1.0\columnwidth]{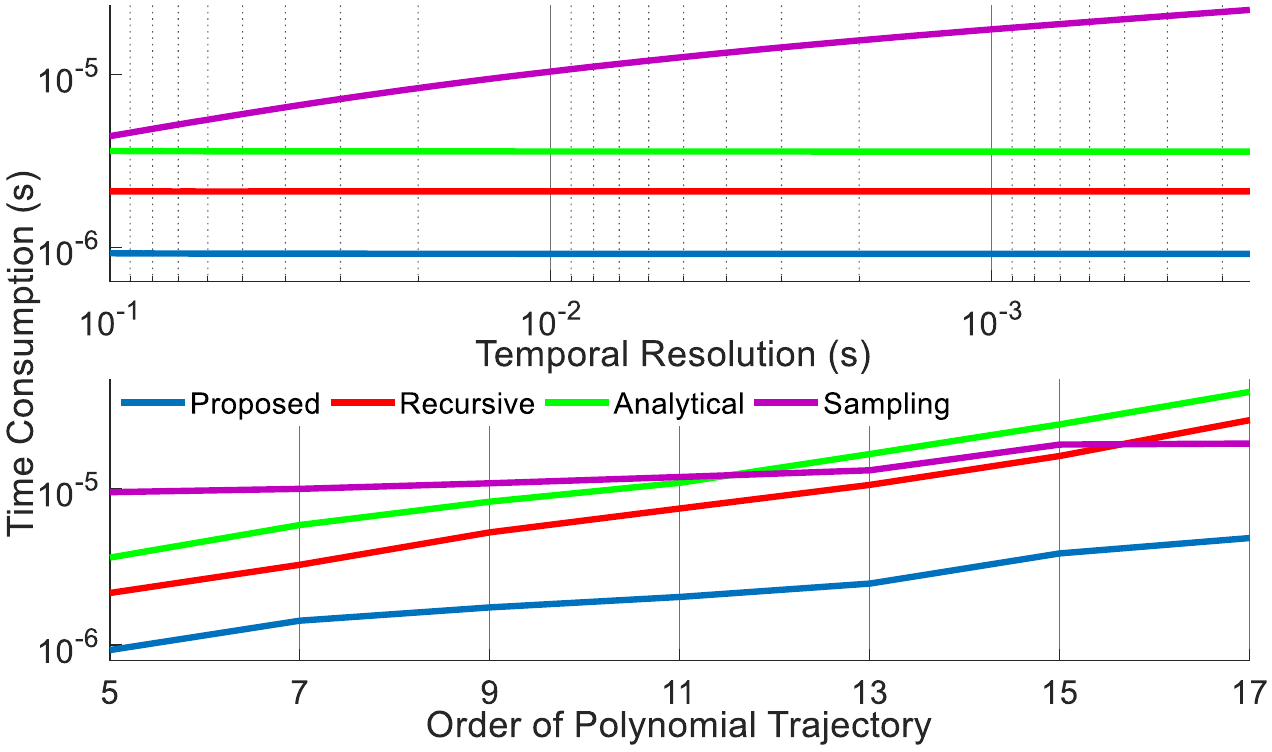}
    \caption{Computation time for feasibility check of speed constraint, under different temporal resolution (upper) and different polynomial order (lower).\label{fig:FeasibilityCheckComparison}}
    \vspace{-0.5cm}
\end{figure}
Firstly, we compare our feasibility check method with Mueller's recursive bound check \cite{Mueller2015ACE}, Burri's analytical extrema check \cite{Burri2015RealtimeVM}, as well as the widely used sampling-based check.
In each case, 1000 trajectory pieces are randomly generated along with velocity constraints to estimate average time consumption. As is shown in Fig.~\ref{fig:FeasibilityCheckComparison}, our method outperforms all other methods in computation speed because of its resolution independence and scalability with higher polynomial orders.
The recursive check and sampling-based check may have false positives under rough temporal resolution.
The efficiency of analytical check and recursive check deteriorates with higher polynomial orders, because both of them involve roots finding which has closed-form solution only for low order ($\leq4$).
In comparison, our method is able to do a solid feasibility check within $1\mu s$.

\subsection{Benchmark for Trajectory Optimization Methods}
\begin{figure}[t]
    \vspace{0.3cm}
    \centering
    \includegraphics[width=1.0\columnwidth]{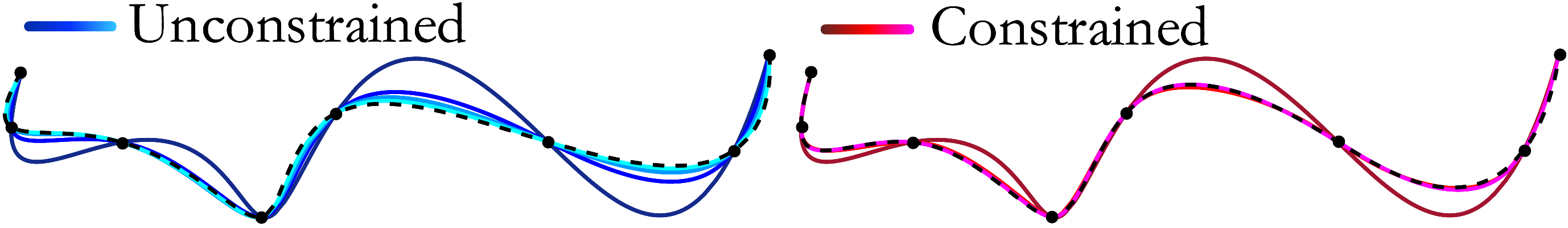}
    \caption{Process of unconstrained/constrained minimization with the same initial guess. Dashed line indicates optimal trajectory in respective case. Constraint on maximum acceleration rate is set to $2.5m/s^2$. The optimized total durations are $10.37s$ (left) and $23.74s$ (right), respectively.\label{fig:TrajOptProcess}}
    \vspace{-0.5cm}
\end{figure}

Secondly, we conduct the benchmark comparison of our trajectory optimization method against state-of-the-art methods.
The benchmark is done as follows: We generate a sequence of waypoints by random walk, of which the step is uniformly distributed over $[-3.0m, 8.0m]$ for each axis.
The maximum speed and acceleration rates are set to $5.0m/s$ and $3.5m/s^2$, respectively. The derivatives on the first and the last waypoints are set to zero.
The objective function is set as $\rho=512.0$, $D_{max}=D_{min}=3$, $w_3=1.0$ and $N=5$.
For a given number of pieces, each method is applied to 1000 sequences of waypoints.
The optimization process stops until the relative decrease of objective is less than $0.001$.
The cost is then normalized by the cost of Alg.~\ref{alg:UnconstrainedSpatialTemporalAM}.
An illustration in Fig.~\ref{fig:TrajOptProcess} shows that the minimized cost in unconstrained case can be used as a baseline.
All comparisons are conducted on an Intel Core i7-8700 CPU under Linux environment.

We compare our method with Richter's method \cite{Richter2013PolynomialTP} and Mellinger's method \cite{Mellinger2011MinimumST}.
Richter's method optimizes trajectory derivatives on waypoints through an unconstrained QP while time allocation is adjusted by gradient descent and scaling.
To use it in constrained case, we soften the constraints by penalizing them in objective function as suggested in \cite{Burri2015RealtimeVM} and directly optimize the time allocation through NLopt \cite{JohnsonNLopt}.
Mellinger's method optimizes the time allocation with total duration fixed first, using backtracking gradient descent (BGD).
Then dynamical feasibility is ensured through time scaling, of which the ratio is properly calculated using Liu's method \cite{Liu2017PlanningDF}.

\begin{figure}[t]
    \begin{center}
        \subfigure[\label{fig:BenchmarkTrajViz} Trajectories in a random map.]
        {\includegraphics[width=0.9\columnwidth]{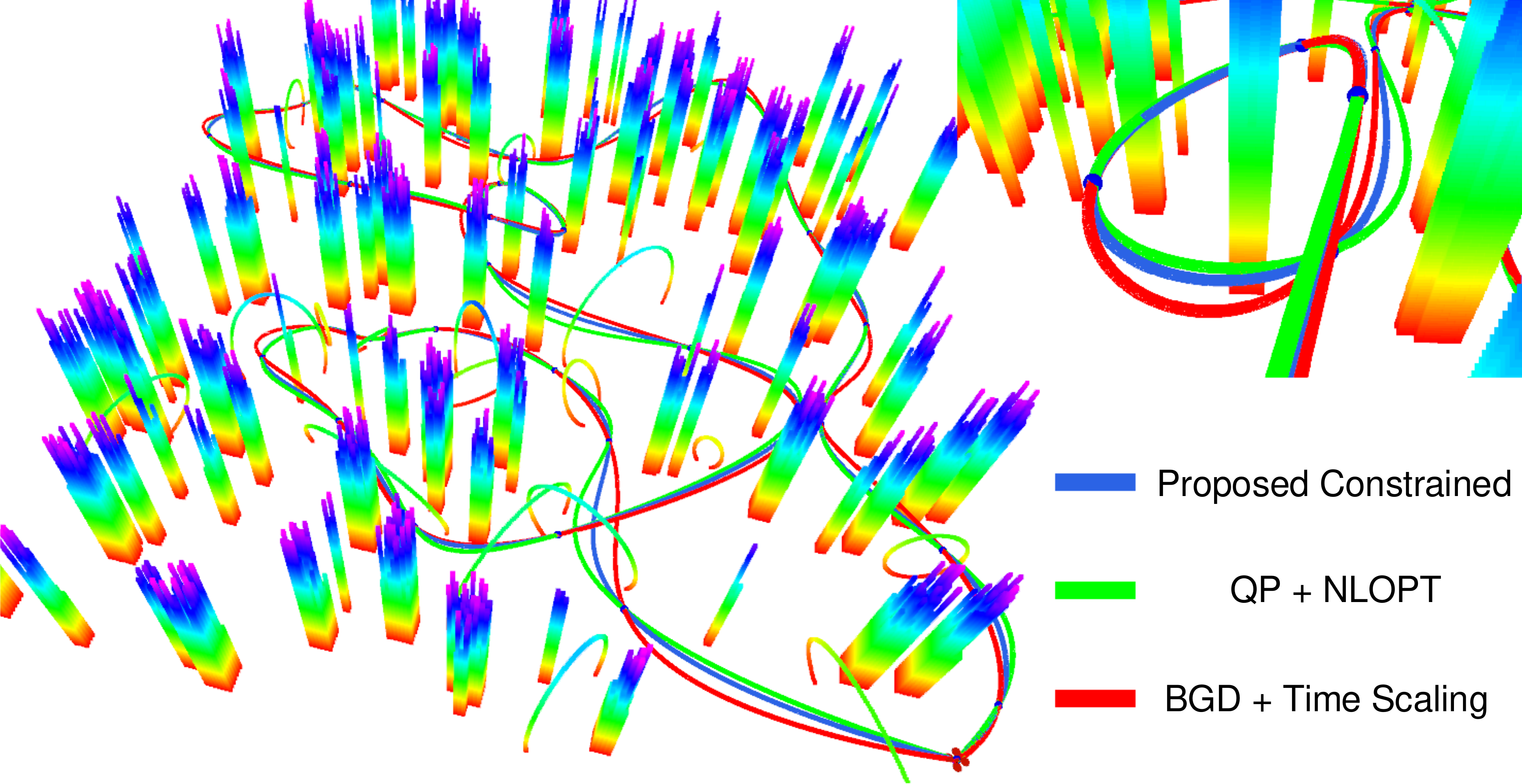}}
        \subfigure[\label{fig:TrajOptBenchmark} Benchmark in computation time (solid) and normalized cost (dotted).]
        {\includegraphics[width=1.0\columnwidth]{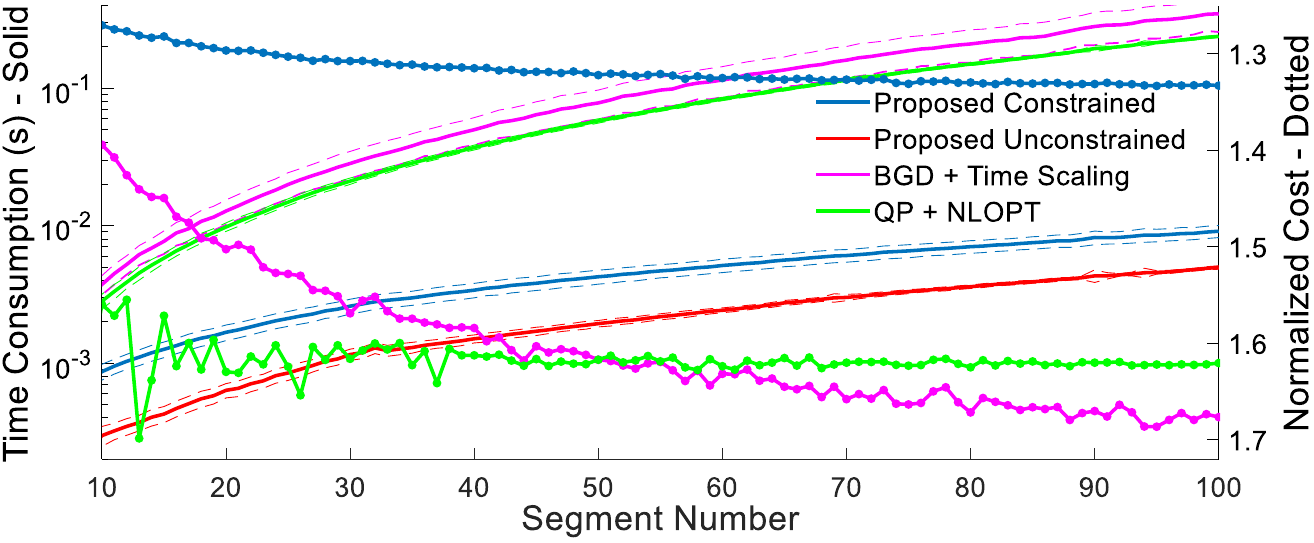}}
    \end{center}
    \vspace{-0.3cm}
    \caption{\label{fig:TrajOptBenchmarkResults} Comparisons of trajectory optimization methods. In Fig.~\ref{fig:BenchmarkTrajViz}, trajectories are generated in a random map with fixed waypoints. Lap times for blue, green and red trajectories are $46.54s$, $62.07s$ and $66.08s$ respectively. In Fig.~\ref{fig:TrajOptBenchmark}, the performance of different methods are provided. Dashed lines indicate standard deviation.}
    \vspace{-0.5cm}
\end{figure}

As is shown in Fig.~\ref{fig:TrajOptBenchmark}, our Alg.~\ref{alg:ConstrainedSpatialTemporalAM} has the fastest speed and the lowest cost when constraints are taken into consideration.
Our method is capable of computing trajectories with $60$ pieces within $5ms$, i.e., $150Hz$ at least.
However, both benchmarked methods fail to accomplish real-time computing for trajectories with more pieces.
Moreover, our Alg.~\ref{alg:ConstrainedSpatialTemporalAM} always obtains better trajectories in terms of the cost function, while benchmarked methods cannot fully utilize the capability of system dynamics.

\subsection{Aggressive Flight Experiment}
\begin{figure}[t]
    \begin{center}
        \subfigure[\label{fig:FlightInstant} Aggressive flight experiment.]
        {\includegraphics[width=1.0\columnwidth]{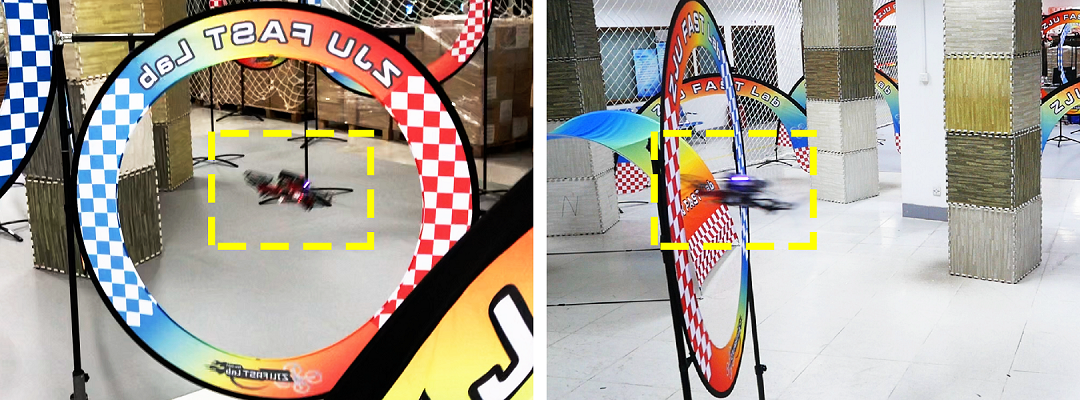}}
        \subfigure[\label{fig:FlightViz} Trajectory of the experiment.]
        {\includegraphics[width=0.9\columnwidth]{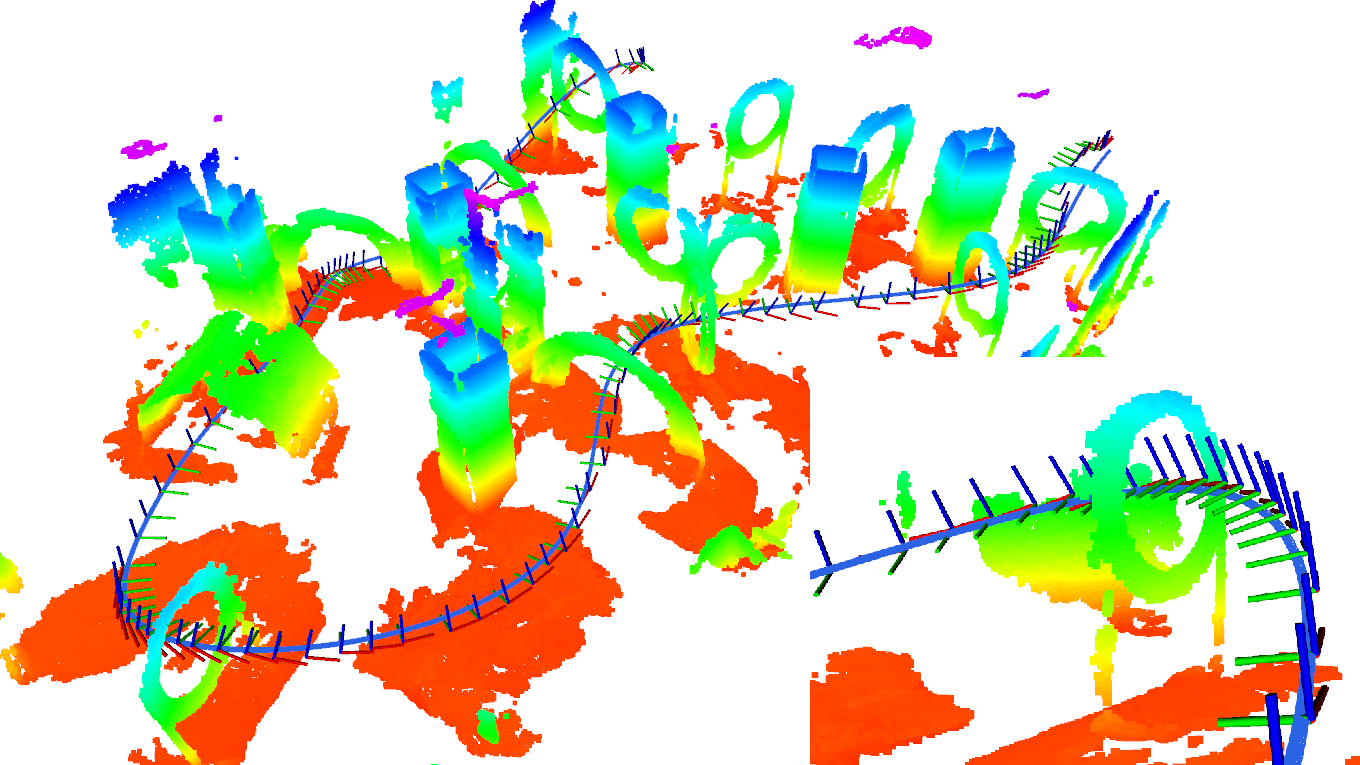}}
        \subfigure[\label{fig:VelAccProfile} Velocity and acceleration against time.]
        {\includegraphics[width=1.0\columnwidth]{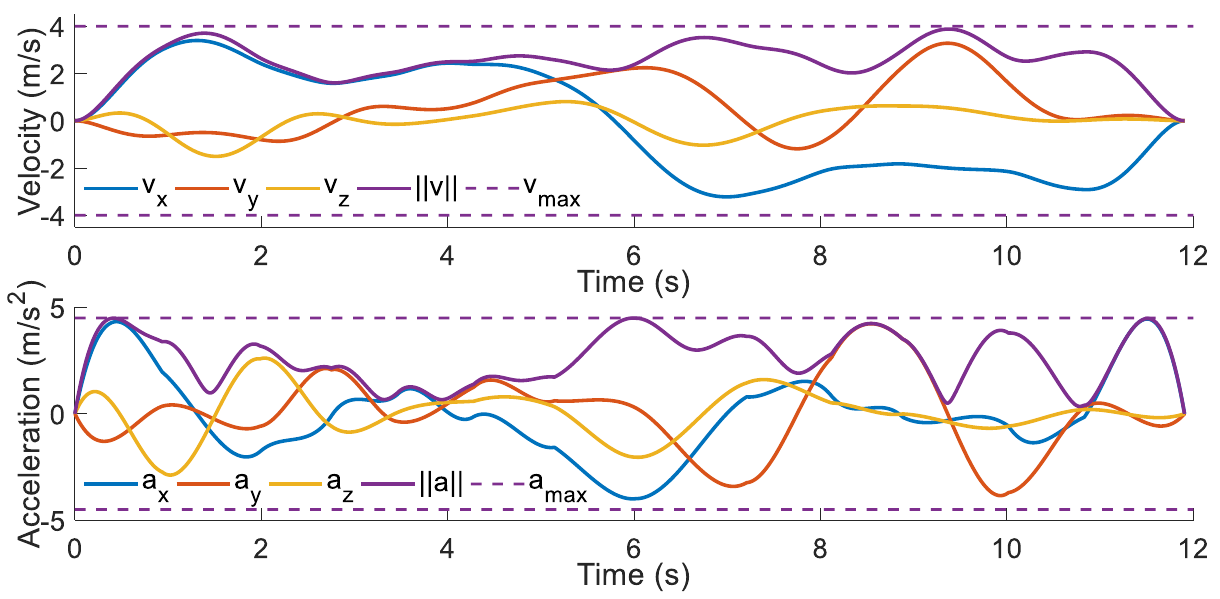}}
    \end{center}
    \vspace{-0.3cm}
    \caption{\label{fig:FinalFlightResults} Details of our aggressive indoor flight. Fig.~\ref{fig:FlightInstant} shows some snapshots of our experiment, where the speed is up to $4.0m/s$. In Fig.~\ref{fig:FlightViz}, the map along with compulsory waypoints is constructed offline. The optimal trajectory is online computed and applied. In Fig.~\ref{fig:VelAccProfile}, velocity/acceleration profiles are provided. The trajectory fully employs capability of the quadrotor in terms of maximum velocity/acceleration rate.}
    \vspace{-0.5cm}
\end{figure}

To validate the performance of our method in real-world applications, we deploy it on a self-developed compact quadrotor platform.
The proposed method is implemented with C++ 11, and all tasks are conducted on an onboard computer with Intel Core i7-8550U CPU.
The pose of our quadrotor is obtained through a robust visual-inertial state estimator \cite{Qin2017VINS}.
Besides, no external positioning system nor offboard computing is used.
A geometric controller is employed for trajectory tracking control \cite{Lee2010GeometricTC}.

The experiment is conducted in a complex indoor scene, which is shown in Fig.~\ref{fig:SequentialInstants}.
A globally consistent map of the scene is pre-built, from which some compulsory waypoints are selected offline.
At the beginning of each flight, an optimal path is produced by RRT* \cite{Karaman2011SamplingbasedAF}, which starts from an initial position and passes all compulsory waypoints.
Our method generates an optimal trajectory online based on this path within milliseconds.
Immediately, the quadrotor starts its aggressive flight.
Different from parameters used in benchmark, we set $v_{max}=4.0m/s$, $a_{max}=4.5m/s^2$ and $\rho=1024.0$.
The aggressive flight along with the generated trajectory is shown in Fig.~\ref{fig:FinalFlightResults}. More details are included in the attached video.

\section{Conclusion}
\label{sec:Conclusion}
In this paper, we propose an efficient trajectory generation method for quadrotor aggressive flight, which has guaranteed convergence and feasibility. Benchmarks for components in our method show its superior computation speed, trajectory quality as well as scalability against state-of-the-art methods.
Aggressive flight experiments in limited space with dense obstacles validate the practical performance of our method. Currently, in the proposed framework, positions of waypoints are fixed during optimization.
However, the method is underlying compatible with waypoints as part of decision variables. Our feasibility checker also supports various safety constraints. In the future, we plan to apply and improve our method in time-critical large-scale motion planning scenarios where complex spatial constraints exist.

\newlength{\bibitemsep}\setlength{\bibitemsep}{0.03\baselineskip}
\newlength{\bibparskip}\setlength{\bibparskip}{0pt}
\let\oldthebibliography\thebibliography
\renewcommand\thebibliography[1]{
    \oldthebibliography{#1}
    \setlength{\parskip}{\bibitemsep}
    \setlength{\itemsep}{\bibparskip}
}
\bibliography{references}

\end{document}